\documentclass[sigconf]{acmart}
\usepackage{amsfonts}
\usepackage{multirow}
\usepackage{verbatim}
\usepackage[font=small]{caption}
\usepackage{cleveref}
\usepackage{enumitem}
\usepackage{makecell}

\newcolumntype{Y}{>{\centering\arraybackslash}X}
\usepackage{amsmath,bm}
\usepackage{subcaption}
\usepackage{graphicx}
\usepackage{tikz}
\usepackage{amsmath}
\usepackage{soul}

\usepackage{booktabs}
\usepackage{colortbl}
\usepackage{multirow}
\usepackage{rotating}

\copyrightyear{2027}
\acmYear{2027}
\setcopyright{cc}
\setcctype{by}
\acmConference[ASPDAC '27]{32nd Asia and South Pacific Design Automation Conference}{January 25--28, 2027}{Tokyo, Japan}
\acmBooktitle{32nd Asia and South Pacific Design Automation Conference (ASPDAC '27), January 25--28, 2027, Tokyo, Japan}
\acmDOI{10.1145/3840404.3848206}
\acmISBN{979-8-4007-2923-2/2027/01}

\begin{document}

\title{FRESHPart: Feedback-Reshaped Spectral Embedding for Hypergraph Partitioning}


\author{Rongjian Liang}
\email{rliang@nvidia.com}
\affiliation{%
  \institution{NVIDIA}
  \city{Austin}
  \state{TX}
  \country{USA}
}

\author{Zhuo Feng}
\email{zfeng12@stevens.edu}
\affiliation{%
  \institution{Stevens Inst of Tech \& NVIDIA}
  \city{Hoboken}
  \state{New Jersey}
  \country{USA}
}

\author{Brucek Khailany}
\email{bkhailany@nvidia.com}
\affiliation{%
  \institution{NVIDIA}
  \city{Austin}
  \state{TX}
  \country{USA}
}

\begin{abstract}
With scalable Laplacian solvers now available, the next frontier is to exploit them more effectively for high-quality hypergraph partitioning. In this work, we revisit spectral hypergraph partitioning from the perspective of spectral embedding geometry. Through a systematic analysis of circuit hypergraphs and controlled synthetic (hyper)graphs, we identify a ray-like embedding structure that helps explain the effectiveness of projection-based partitioning. We further show that the geometry of spectral embeddings can be systematically influenced through Laplacian design and hyperedge reweighting. Guided by these insights, we propose FRESHPart, a feedback-reshaped spectral partitioning framework that refines the spectral embedding through Geometry Feedback and Search Feedback, producing embeddings that are better aligned with projection-based partitioning. On the Titan23 benchmark, FRESHPart achieves average cut reductions over hMETIS of 13.09\% at $\epsilon=2\%$ and 25.21\% at $\epsilon=20\%$, surpassing the strongest previously reported results (12.16\% and 22.53\%, respectively), while completing every benchmark in under 120 s on a single GPU.

\end{abstract}







\maketitle

\section{Introduction}
Hypergraph partitioning~\cite{bretto2013hypergraph, karypis1997multilevel} is a fundamental optimization problem in VLSI physical design and many scientific computing applications. Given a hypergraph, the objective is to partition its vertices into balanced subsets while minimizing the number (or weight) of cut hyperedges. Scalable and high-quality partitioning algorithms are essential for modern large-scale applications.

\begin{figure}[!t]
\centering
\includegraphics[width=0.97\linewidth]{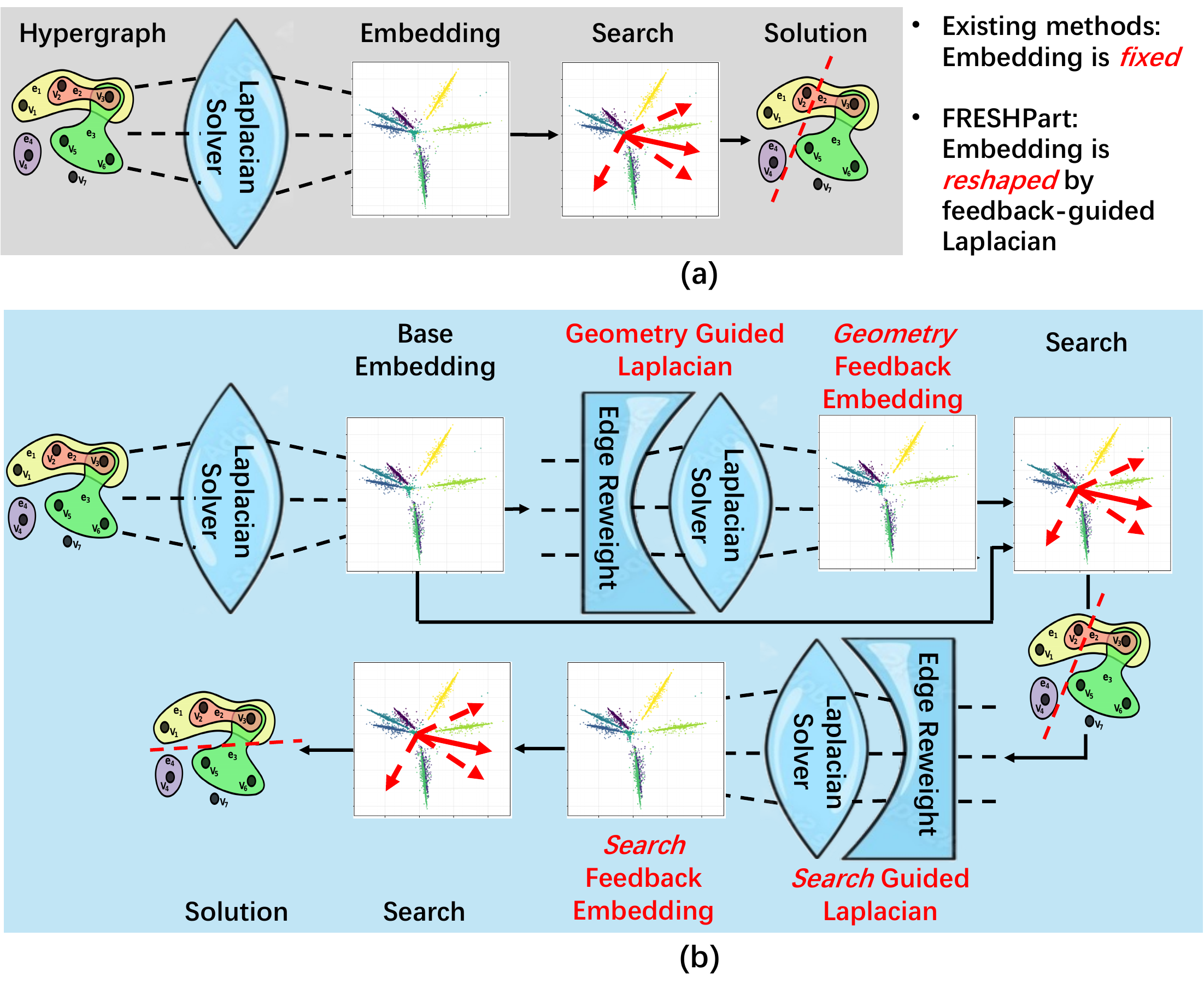}%
\vspace{-5mm}%
\caption{Comparison of (a) workflow of prior spectral partitioners and (b) workflow of FRESHPart.}
\vspace{-6mm}
\label{fig:workflow}
\end{figure}

Spectral graph partitioning~\cite{fiedler1975property, shi2000normalized, hagen1992new, pothen1990partitioning, ding2001min,zien1999multilevel,alpert1999spectral} has been studied for decades due to its strong theoretical foundations, but its practical adoption has been limited by two challenges: the high cost of computing spectral embeddings and partition quality inferior to the dominant multilevel paradigm~\cite{karypis1998hmetis,schlag2016k,sajadinia2025shypar,lee2025hyperg,chen2026genpart}. Recent advances in Laplacian solvers~\cite{spielman2004nearly, liu2024efficient,gottesburen2024scalable,chan2018spectral} have largely removed the computational bottleneck, making spectral methods practical for million-scale sparse graphs. Consequently, \ul{the primary challenge has shifted from computing spectral embeddings efficiently to exp- loiting them effectively}. This motivates a fresh look at spectral hypergraph partitioning with the goal of surpassing state-of-the-art (SOTA) multilevel partitioners.

Classical spectral partitioning typically clusters spectral embeddings using $K$-means, which favors compact Euclidean clusters and provides no direct mechanism for enforcing balance constraints~\cite{von2007tutorial, ng2001spectral}. More recently, projection-based partitioning, exemplified by HySpecPro~\cite{liang2026hyspecpro}, has emerged as an effective alternative by directly searching for projection vectors in the spectral embedding space that optimize both cut quality and balance. Its success, however, has been supported largely by empirical evidence, raising two fundamental questions: \textit{What geometric properties of embeddings make projection-based partitioning effective}, and \textit{what determines the geometry of spectral embeddings?} To answer these questions, we systematically analyze spectral embeddings of both circuit hypergraphs and controlled synthetic (hyper)graphs using principal component analysis (PCA) visualization and a new ray concentration metric. Our study reveals that the ray structure provides a geometric explanation for the effectiveness of projection-based partitioning. Moreover, we show that the geometry is determined not only by the structural properties of the underlying (hyper)graph, but also by the choice of Laplacian operator and hyperedge weighting. These findings demonstrate that \ul{spectral geometry is not merely an intrinsic property of the input hypergraph, but can be systematically shaped to improve partitioning quality}.

Motivated by these insights, we propose FRESHPart, a feedback-driven spectral partitioning framework that treats the Laplacian as an adaptive lens for generating spectral embeddings. As illustrated in \Cref{fig:workflow}, conventional spectral methods compute the Laplacian once and use it as a fixed lens throughout partitioning. If this lens is not well aligned with the partitioning objective, additional projection search can only explore the resulting embedding more thoroughly—it cannot correct the misalignment introduced by the lens itself. FRESHPart instead leverages both Geometry Feedback from the current embedding and Search Feedback from partitioning outcomes to reweight hyperedges and reshape the Laplacian, producing embeddings that are better aligned with the partitioning objective. Although this refinement requires additional Laplacian solves and projection searches, it preserves the linear runtime scalability of HySpecPro while delivering substantially larger gains than simply increasing the projection search budget.

Our main contributions are summarized as follows:
\begin{itemize}[topsep=2pt, leftmargin=4.5mm]
  \item \ul{We provide a geometric perspective on projection-based spectral hypergraph partitioning}. Through systematic analysis of circuit hypergraphs and controlled synthetic (hyper)graphs, we identify a ray-like embedding structure that helps explain the effectiveness of projection-based partitioning, and show how graph structure, Laplacian design, and hyperedge reweighting influence spectral geometry.
  
  \item \ul{We propose FRESHPart, a feedback-driven spectral hypergraph partitioning framework} that reshapes the spectral embeddings according to both embedding geometry and partitioning outcomes. By reweighting hyperedges and recomputing the bipartite Laplacian, FRESHPart produces embeddings that are better aligned with the partitioning objective.
  
  \item \ul{We demonstrate SOTA partitioning quality on the Titan23 bench- mark suite}. FRESHPart consistently outperforms existing multilevel and spectral partitioners, achieving the best reported cut quality while completing every benchmark in under 120 seconds on a single GPU.
\end{itemize}

\section{Preliminaries}
\label{sec:related}
\subsection{Problem Formulation}
A hypergraph is defined as $H=(V,E)$, where $V$ is the vertex set and $E$ is the set of hyperedges, with each hyperedge $e\subseteq V$ connecting an arbitrary subset of vertices. Each vertex $v\in V$ and hyperedge $e\in E$ is associated with a nonnegative weight $w(v)$ and $w(e)$, respectively. The balanced hypergraph partitioning problem seeks to partition $V$ into $K$ disjoint blocks $\{V_1,\ldots,V_K\}$ while minimizing the total weight of cut hyperedges subject to
\[
\max_{1\le k\le K}\sum_{v\in V_k}w(v)
\le
\left(\frac{1}{K}+\epsilon\right)\sum_{v\in V}w(v),
\]
where $\epsilon$ specifies the allowable imbalance ratio. This work focuses on the bipartitioning problem ($K=2$); extending the proposed framework to multi-way partitioning is left for future work.

\subsection{Spectral Embeddings and Partitioning}
Spectral partitioning relaxes the discrete partitioning problem into a continuous one by embedding vertices into a low-dimensional Euclidean space. The embedding is obtained from the eigenvectors associated with the smallest nontrivial eigenvalues of a normalized Laplacian, providing a continuous relaxation of the balanced minimum-cut problem~\cite{von2007tutorial}.

Two Laplacian formulations~\cite{zhou2006learning} are widely used for hypergraphs. The \emph{clique expansion} replaces each hyperedge $e$ with a weighted clique connecting all pairs of its incident vertices. Each pairwise edge is assigned weight $w(e)/\binom{|e|}{2}$, where $|e|$ denotes the degree of the hyperedge. Let $A_C$ and $D_C$ denote the resulting weighted adjacency and degree matrices. The normalized Laplacian is $L_C=I-D_C^{-1/2}A_CD_C^{-1/2}$.
Although simple and computationally efficient, clique expansion reduces higher-order interactions to pairwise connections and may distort the original higher-order connectivity, particularly for large hyperedges.

The \emph{bipartite representation} instead preserves the incidence structure by introducing one node for each hyperedge. A bipartite edge connects a vertex $v$ to a hyperedge node $e$ with weight $w(e)/|e|$ whenever $v\in e$. Let $A_B$ and $D_B$ denote the resulting adjacency and degree matrices. The normalized bipartite Laplacian is $L_B=I-D_B^{-1/2}A_BD_B^{-1/2}$.

Classical spectral partitioning obtains partitions by clustering the embedded vertices using \emph{K}-means. More recently, HySpecPro employs Covariance Matrix Adaptation Evolution Strategy (CMA-ES~\cite{hansen2019pycma}) to optimize projection vectors in the spectral embedding space, assigning each vertex to the partition whose projection vector has the largest inner product. This projection-based framework directly optimizes cut quality under balance constraints while achieving linear runtime in the total hyperedge degree. Building on this framework, we investigate the geometry of spectral embeddings and propose a feedback mechanism that reshapes the embedding through hyperedge reweighting and Laplacian recomputation.

\vspace{-2mm}
\section{Spectral Geometry Diagnostics}
\label{sec:synthetic-geometry}
We systematically investigate the geometry of spectral embeddings through two complementary studies. We first analyze the Titan23 benchmarks to understand why projection-based partitioning is effective. We then study controlled synthetic (hyper)graphs to identify what determines the geometry of spectral embeddings.

\vspace{-3mm}
\subsection{Diagnostic Tools}
Two complementary diagnostic tools are introduced: \emph{PCA visualization} for qualitative inspection and a new \emph{ray concentration metric} for quantitative analysis.

\textbf{PCA Visualization.}
We construct spectral embeddings using both the normalized bipartite and clique Laplacians. After removing numerically trivial eigenmodes (eigenvalues below $10^{-6}$), we retain the first $S=16$ nontrivial eigenvectors $u_1,\ldots,u_S$. The embedding of vertex $v$ is defined as $x_v=\bigl(u_1(v),\ldots,u_S(v)\bigr)$. For visualization, the embeddings are projected onto two dimensions using PCA.

\textbf{Ray Concentration.}
To quantify ray structure, we apply spherical $K$-means to the embedding.
We first normalize each vertex embedding as
$\hat{x}_v = x_v / \|x_v\|_2$, then cluster the normalized vectors into
$K_r$ groups with unit-norm centers $c_1,\ldots,c_{K_r}$. We define
\emph{ray concentration} as
\[r_{K_r}(X)
=
\frac{1}{|V|}
\sum_{v\in V}
\max_{1\le j\le K_r}
\hat{x}_v^\top c_j.\]
This measures the average cosine similarity to the nearest ray center.
Higher values indicate that the embeddings align closely with a small
number of rays.

For the Titan23 benchmarks, we fix $K_r=12$, which captures the dominant ray structure without introducing overly fine-grained ray clusters. For the synthetic benchmarks, we fix $K_r=8$, matching the eight underlying modules in the synthetic graph generation.

\begin{figure}[!t]
\centering
\vspace{-4mm}
\includegraphics[width=1.0\linewidth]{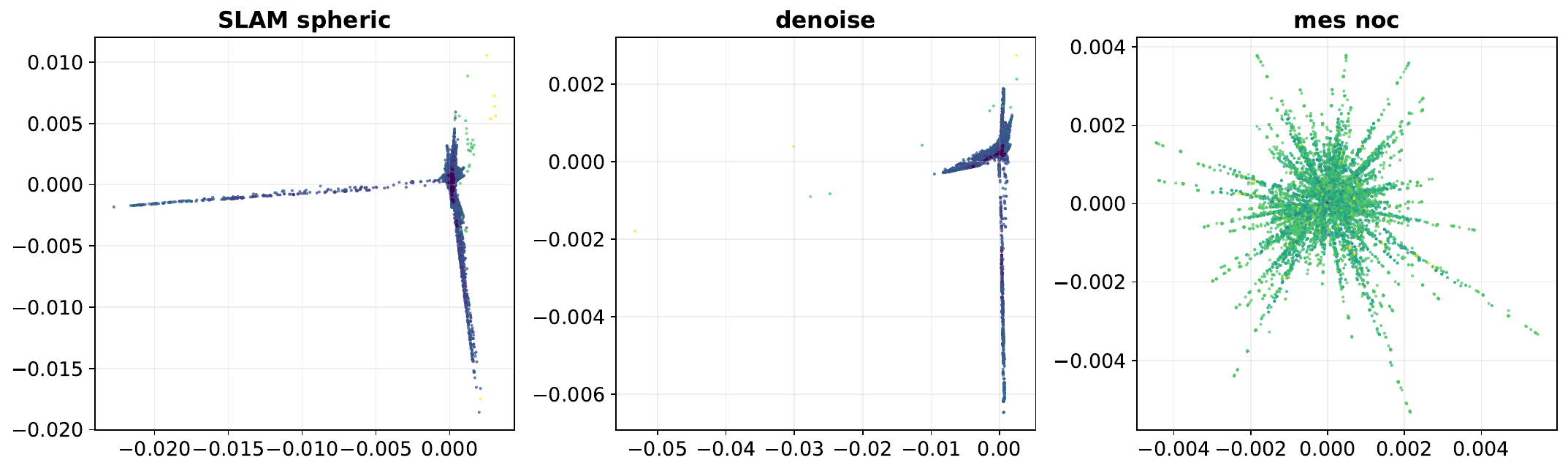}%
\vspace{-3mm}%
\caption{PCA visualization of bipartite spectral embeddings on representative Titan23 benchmarks.}
\vspace{-2mm}
\label{fig:titan23}
\end{figure}

\begin{figure}[!t]
\centering
\vspace{-2.5mm}
\includegraphics[width=1.0\linewidth]{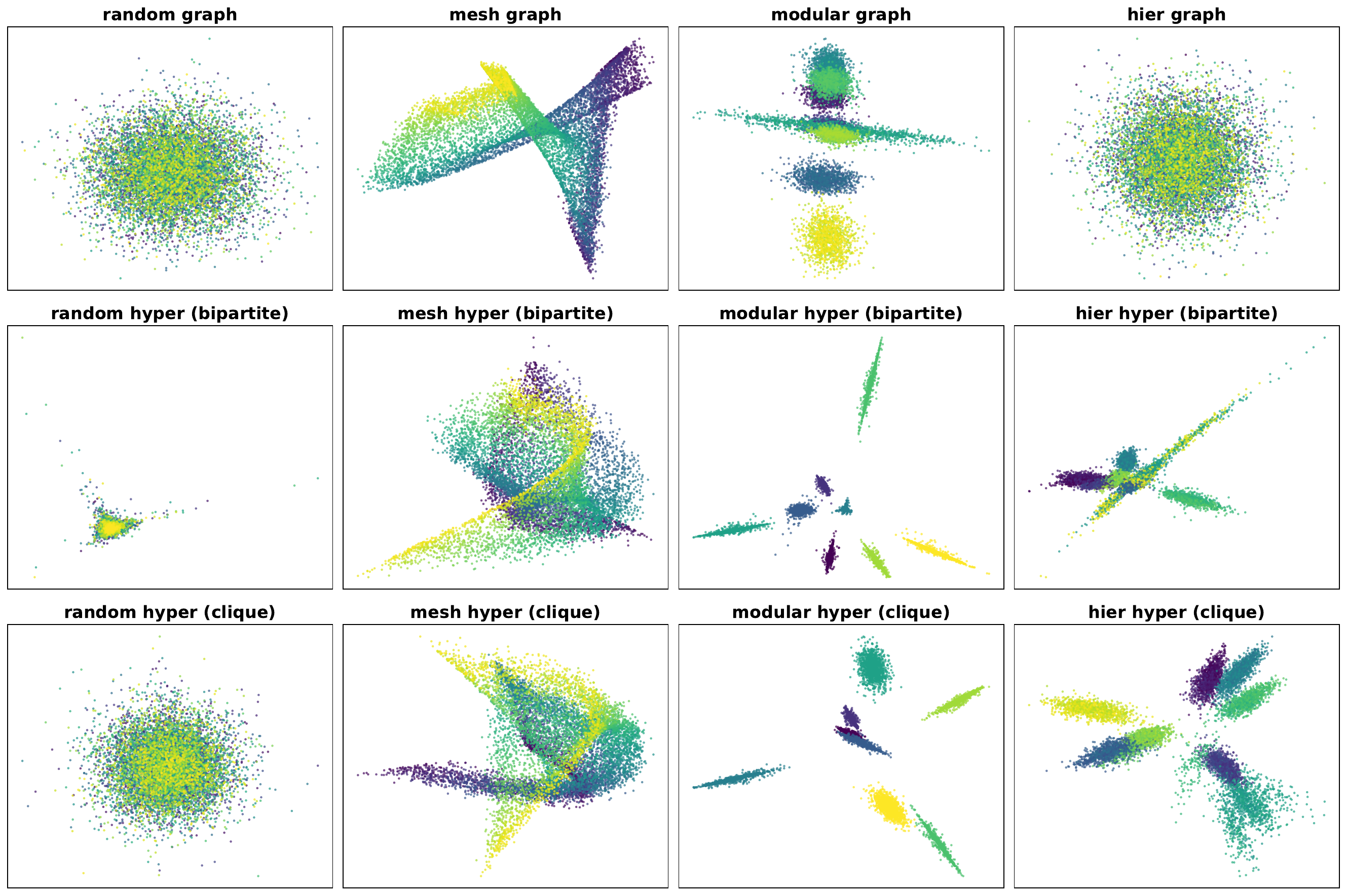}%
\vspace{-3mm}%
\caption{PCA visualization of spectral embeddings for synthetic graph and hypergraph families at average vertex degree $\bar d=20$. Hypergraph embeddings are computed using both the normalized bipartite and clique Laplacians. Colors indicate the ground-truth modules used to generate the synthetic instances.}
\vspace{-6mm}
\label{fig:synthetic}
\end{figure}

\subsection{Diagnostic of Titan23 Benchmarks}
The Titan23 benchmark suite consists of real-world VLSI circuit hypergraphs. As shown in \Cref{fig:titan23}, their bipartite spectral embeddings exhibit a pronounced ray-like structure, rather than the compact blob-like distributions typically assumed by classical $K$-means clustering. This observation is further quantified in \Cref{tab:evolution-k12}: the base embeddings achieve an average ray concentration of 0.922, indicating that \ul{ray-like spectral geometry is prevalent in real-world circuit hypergraphs}.

To examine whether the ray structure is associated with the effectiveness of projection-based partitioning, we compute the Pearson correlation between the ray concentration and the cut-size reduction achieved by HySpecPro over the widely used multilevel partitioner hMetis at $\epsilon=2\%$. The resulting correlation coefficient of 0.42 indicates a moderate positive correlation, suggesting that \ul{benchmarks with stronger ray structure tend to benefit more from projection-based partitioning}. The correlation is not expected to be perfect, as partition quality is also influenced by factors beyond spectral geometry.

\subsection{Diagnostic of Controlled Synthetic Graphs}

To understand what determines the geometry of spectral embeddings, we study controlled synthetic (hyper)graph families. The goal is to isolate the effects of graph structure and Laplacian design on the ray structure observed in the Titan23 benchmarks.

\begin{table}[!t]
\centering
\scriptsize
\setlength{\tabcolsep}{3pt}
\vspace{-4mm}
\caption{Ray concentration across synthetic graph families at different average vertex degrees. Hypergraph results are reported for both the bipartite (bi.) and clique (cl.) Laplacians.}
\vspace{-4mm}
\label{tab:synthetic-ray-degree}
\begin{tabular}{l|ccc}
\toprule
\textbf{Family/operator} & $\bar d\!\approx\!1.25$ & $\bar d\!\approx\!5$ & $\bar d\!\approx\!20$ \\
\midrule
random graph              & 0.802 & 0.754 & 0.567 \\
random hyper (bi.)        & 0.833 & 0.806 & 0.624 \\
random hyper (cl.)        & 0.866 & 0.751 & 0.566 \\
\midrule
mesh graph                & 0.759 & 0.826 & 0.826 \\
mesh hyper (bi.)          & 0.814 & 0.825 & 0.825 \\
mesh hyper (cl.)          & 0.830 & 0.823 & 0.826 \\
\midrule
modular graph             & 0.802 & 0.941 & 0.910 \\
modular hyper (bi.)       & 0.836 & 0.952 & 0.925 \\
modular hyper (cl.)       & 0.886 & 0.943 & 0.872 \\
\midrule
hierarchical graph        & 0.745 & 0.731 & 0.567 \\
hierarchical hyper (bi.)  & 0.840 & 0.925 & 0.955 \\
hierarchical hyper (cl.)  & 0.821 & 0.926 & 0.947 \\
\bottomrule
\end{tabular}
\vspace{-2mm}
\end{table}

\begin{table}[!t]
\centering
\scriptsize
\setlength{\tabcolsep}{3pt}
\vspace{-2mm}
\caption{Effect of hyperedge weighting
$|e|^{-\alpha}$ on the ray concentration of modular hypergraphs under the bipartite Laplacian.}
\vspace{-4mm}
\label{tab:modular-hyper-alpha}
\begin{tabular}{c|cccccc}
\toprule
$\bar d$ & $\alpha{=}0$ & $0.5$ & $0.6$ & $1.0$ & $1.5$ & $2.0$ \\
\midrule
1.25 & 0.860 & 0.883 & \textbf{0.893} & 0.836 & 0.861 & 0.873 \\
5    & 0.943 & \textbf{0.954} & \textbf{0.954} & 0.952 & 0.918 & 0.927 \\
20   & 0.864 & 0.883 & 0.884 & \textbf{0.925} & 0.914 & 0.892 \\
\bottomrule
\end{tabular}
\vspace{-6mm}
\end{table}

\textbf{Controlled synthetic (hyper)graphs.}
We construct four graph families and their hypergraph counterparts to represent distinct connectivity patterns: random graphs with no global organization, mesh graphs with local geometric connectivity, modular graphs with community structure, and hierarchical graphs with multi-scale organization. All instances contain $10{,}000$ vertices. The graph families are generated using global random edges, local grid edges, an eight-block stochastic block model with $98\%$ within-block and $2\%$ cross-block edges, and an $8$-ary recursive partition tree with leaf size at most $16$, respectively.
The corresponding hypergraphs preserve the same structural patterns through analogous hyperedge sampling. Hyperedge sizes are sampled uniformly from $\{2,\ldots,6\}$, yielding an average size of approximately $4$ pins, consistent with the Titan23 benchmarks.

To enable fair comparisons, we evaluate three average vertex degrees, $\bar d\in\{1.25,5,20\}$, over ten random seeds. For graphs,
$\bar d=\frac{2|E|}{|V|}$,
whereas for hypergraphs we use the average vertex incidence degree,
$\bar d=\frac{1}{|V|}\sum_{e\in E}|e|$.

\textbf{Main empirical findings.}
Figure~\ref{fig:synthetic} and Table~\ref{tab:synthetic-ray-degree} reveal three key observations. First, \ul{hypergraphs consistently exhibit stronger ray-like embeddings than their graph counterparts at the same density}, particularly for structured families.

Second, \ul{modular and hierarchical connectivity is an important source of persistent ray structure}. Although sparse random graphs may also exhibit ray-like embeddings, this effect largely disappears as density increases. In contrast, modular and hierarchical hypergraphs maintain high ray concentration even at realistic densities, suggesting that the observed geometry arises from coherent graph structure rather than sparsity alone.

Third, \ul{the choice of Laplacian operator significantly influences embedding geometry}. The bipartite Laplacian consistently preserves stronger ray structure than the clique-expansion Laplacian, consistent with the superior performance of projection-based partitioning under the bipartite formulation reported in~\cite{liang2026hyspecpro}.

Finally, we investigate the effect of hyperedge weighting by replacing the standard bipartite normalization, $w(e)/|e|$, with the generalized weighting scheme $w(e)/|e|^{\alpha}$, where $\alpha\in\{0,0.5,0.6,1,1.5,2\}$. As shown in Table~\ref{tab:modular-hyper-alpha}, moderate normalization strengthens the ray geometry, whereas overly aggressive normalization weakens it. This demonstrates that \ul{hyperedge reweighting provides an effective mechanism for influencing spectral geometry} and motivates the feedback strategies introduced in the next section.
\section{Feedback-Reshaped Embeddings for Hypergraph Partitioning}
\label{sec:dynamic}
\Cref{fig:workflow}(b) illustrates the FRESHPart workflow. Guided by
the geometric analysis in \Cref{sec:synthetic-geometry}, we first
compute a base embedding from the bipartite Laplacian. Geometry
Feedback clusters this embedding into rays and reweights each
hyperedge according to the number of rays it spans. Recomputing the Laplacian yields a geometry-guided embedding.
CMA-ES projection search on the base and geometry-guided embeddings
then produces diverse, high-quality candidate partitions. Search
Feedback uses their hyperedge cut frequencies together with the ray
spans to construct a new set of weights. A further Laplacian solve
yields the search-guided embedding for the second round of projection
search. Reweighting changes only the embeddings used for search;
candidate cuts and balance are evaluated on the original hypergraph.
Before each search round, lightweight probes guide the allocation of
the CMA-ES budget across candidate embeddings. The following
subsections detail the two feedback signals and budget allocation.

\vspace{-2mm}
\subsection{Geometry Feedback}
\label{sec:dyn-ray}

Motivated by the positive correlation between ray concentration and partitioning quality, FRESHPart reshapes the spectral geometry by emphasizing hyperedges that reinforce ray structure while suppressing those that blur it. 

Specifically, it first clusters the base embedding into rays using spherical $K$-means described in \Cref{sec:synthetic-geometry}, and then counts the number of distinct rays $R(e)$ spanned by each hyperedge $e\in E$. A simple monotonic cross-ray factor
\begin{equation*}
    w_g(e)= b(e)\,R(e)^{-\beta}
\end{equation*}
is adopted, as visualized in \Cref{fig:geo}.
The term $R(e)^{-\beta}$ suppresses hyperedges spanning multiple rays, as they tend to blur the ray geometry. Conversely, $b(e)=\rho>1$ if all vertices of $e$ lie on the same ray, and $b(e)=1$ otherwise, providing a mild boost to ray-consistent hyperedges. We use $\beta=1.0$ and $\rho=1.2$. The cross-ray factor $w_g(e)$ is mean-normalized for numerical stability. The final Geometry Feedback weight is then obtained by multiplying the normalized cross-ray factor with the standard bipartite normalization $w(e)/|e|$. Overall, the reweighting strengthens compact, ray-consistent hyperedges while suppressing cross-ray hyperedges, producing spectral embeddings with sharper ray geometry for projection-based search.

\vspace{-2mm}
\subsection{Search Feedback}

While Geometry Feedback relies solely on the spectral embedding, Search Feedback closes the optimization loop by allowing the partitioning result to reshape the next embedding. Instead of relying on a single partition solution, FRESHPart collects multiple \emph{diverse} high-quality partitions. Hyperedges that are consistently preserved across these solutions likely represent stable connectivity, whereas those that repeatedly appear on cut boundaries reveal recurring separation regions. FRESHPart leverages this information to reweight hyperedges and construct a refined embedding that is better aligned with the partitioning objective.

Let $\mathcal{S}=\{s_1,\ldots,s_m\}$ denote the set of candidate bipartitions returned by the first round of projection search. To avoid collecting nearly identical solutions, we iteratively select the best partitions while enforcing a minimum Hamming distance between them, ensuring that $\mathcal{S}$ captures diverse high-quality cuts. We then define the cut frequency of a hyperedge $e$ as
\begin{equation*}
    f(e)=\frac{1}{m}\sum_{i=1}^{m}
\mathbf{1}\{e \text{ is cut by } s_i\},
\end{equation*}
which measures how frequently $e$ appears on the boundaries of promising partitions. Based on the cut frequency and the ray span $R(e)$ computed from the base embedding, FRESHPart computes the cut-history factor
\[
 w_{s}(e)=
\begin{cases}
1, & f(e)\ge \tau_f \text{ and } R(e)<\tau_\sigma,\\
\exp\!\left\{-\alpha\left(f(e)-\frac{1}{2}\right)\right\},
& \text{otherwise},
\end{cases}
\]
followed by mean normalization and clipping for numerical stability. 
The final Search Feedback weight is the multiplication of the cut-history factor and the standard bipartite normalization.

\begin{figure}[!t]
\centering
\vspace{-4mm}
\begin{minipage}[t]{0.495\linewidth}
    \centering
    \includegraphics[width=\linewidth]{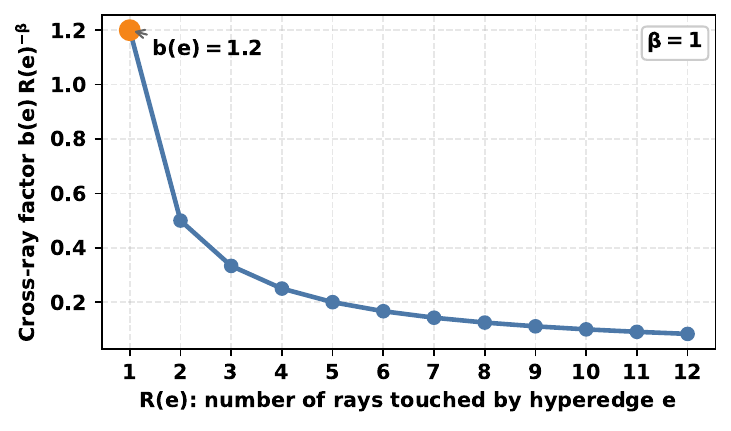}
    \vspace{-8mm}
    \caption{Cross-ray factor.}
    \label{fig:geo}
\end{minipage}
\hfill
\begin{minipage}[t]{0.495\linewidth}
    \centering
    \includegraphics[width=\linewidth]{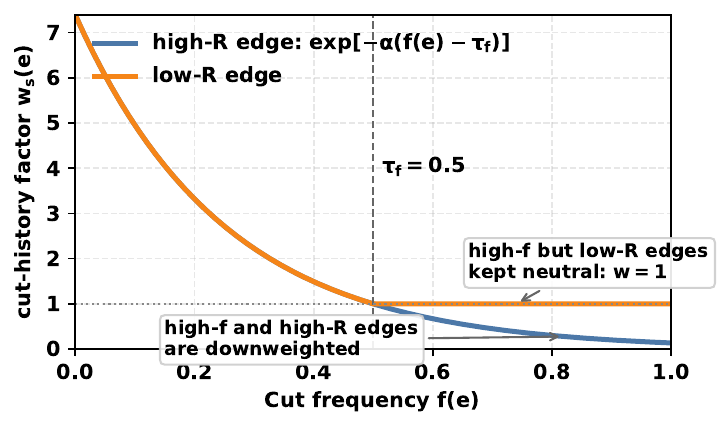}
    \vspace{-8mm}
    \caption{Cut-history factor.}
    \label{fig:cut}
\end{minipage}
\vspace{-4mm}
\end{figure}

As depicted in \Cref{fig:cut}, this rule strengthens hyperedges that are consistently preserved, encouraging their incident vertices to remain compact in the refined embedding. Hyperedges that are both frequently cut and span many rays are instead downweighted, as they are more likely to blur the embedding and interfere with high-quality partitions. In contrast, frequently cut hyperedges that remain geometrically compact are kept at neutral weight, since they may correspond to meaningful local boundaries rather than noisy global constraints.

In our implementation, we set $\alpha=4$, $\tau_f=0.5$, and $\tau_\sigma$ to the median ray span among high-frequency hyperedges, and clip the normalized weights to $[0.2,5.0]$. Together with Geometry Feedback, Search Feedback enables FRESHPart to iteratively reshape the spectral embedding using both its geometric structure and information revealed by the partition search.

\vspace{-2mm}
\subsection{Adaptive Search Budget Allocation}

FRESHPart adaptively allocates the CMA-ES search budget across candidate embeddings using inexpensive probe searches, improving runtime efficiency while maintaining partition quality.

For each Laplacian solve shown in \Cref{fig:workflow}(b), we compute the first $36$ nontrivial eigenvectors and generate four candidate embeddings by retaining the first $8$, $16$, $24$, and $36$ dimensions, respectively. These embeddings provide multiple search spaces with different spectral resolutions, which we empirically find helps projection search escape local optima.

Before each round of CMA-ES search, FRESHPart performs a lightweight probe search to estimate the potential of each embedding. For every candidate embedding $X_j$, we evaluate a small set of random projection directions and record the best cut-and-balance objective value $q_j$. Let
$q_{\min}=\min_j q_j,$ and $\rho_j=\frac{q_j}{q_{\min}}$,
where a smaller $\rho_j$ indicates a more promising embedding.

For the base and Geometry Feedback embeddings, candidates with $\rho_j\le1.3$ receive the full CMA-ES budget, those with $1.3<\rho_j\le2.0$ receive a reduced budget, and the remaining candidates are discarded. Because Search Feedback uses partitions from
the first search round, we instead compare its probe result with
the best solution from that round.

In our implementation, a full budget consists of five CMA-ES restarts, whereas a reduced budget uses at most three restarts. Each restart runs for at most 60 iterations with a population size of 2000, and terminates early if no improvement is observed for 5 consecutive iterations. This adaptive policy concentrates computation on the most promising embeddings.

\vspace{-1.5mm}
\subsection{Runtime Complexity}
HySpecPro~\cite{liang2026hyspecpro} establishes that both spectral embedding construction and CMA-ES-based projection search scale linearly with the total hyperedge degree (TotalDeg), yielding an overall runtime of $O(\text{TotalDeg})$ when the embedding dimension, CMA-ES population size, and optimization iterations are treated as constants.

Compared with HySpecPro, FRESHPart introduces two additional feedback stages. Geometry Feedback clusters the spectral embeddings using spherical $K$-means, computes the ray span of each hyperedge, and updates the hyperedge weights. Search Feedback computes hyperedge cut frequencies from a small set of high-quality partition solutions and performs another linear-time hyperedge reweighting pass. These operations require only linear passes over the embedding vectors and hyperedge incidences.

FRESHPart performs a constant number of Laplacian solves to construct the base, Geometry Feedback, and Search Feedback embeddings. Although the CMA-ES search budget is adaptively allocated across candidate embeddings, the total search budget is bounded by a fixed constant independent of the hypergraph size. Therefore, the additional feedback affects only the constant factor, and the overall runtime of FRESHPart remains $O(\text{TotalDeg})$, preserving the linear scalability of HySpecPro.
\vspace{-2.5mm}
\section{Experimental Results and Analysis}
\label{sec:exp}
FRESHPart is implemented in Python. Spectral embedding construction is accelerated using CuPy~\cite{nishino2017cupy}. Projection search uses the CMA-ES library~\cite{hansen2019pycma}, with batched solution evaluation accelerated by the Deep Graph Library~\cite{wang2019deep}. The Geometry and Search Feedback computations are implemented using custom Python code. All experiments are conducted on a server with AMD EPYC 7742 CPUs and an NVIDIA V100 GPU. Our
evaluation suite is the $22$ Titan23 benchmarks, whose characteristics are shown in \Cref{tab:evolution-k12}, run at both the tight
($\epsilon{=}2\%$) and relaxed ($\epsilon{=}20\%$) balance constraints. We compare FRESHPart against the following SOTA partitioners:
\begin{itemize}[topsep=2pt, leftmargin=4.5mm]
    \item \textbf{hMETIS}~\cite{karypis1998hmetis}: a widely used sequential multilevel partitioner;
    \item \textbf{SpecPart}~\cite{bustany2022specpart}: a refinement method that improves hMETIS using spectral embeddings;
    \item \textbf{MedPart}~\cite{liang2024medpart}: a GPU-accelerated multilevel evolutionary differentiable partitioner;
    \item \textbf{KaHyPar}~\cite{schlag2016k}: a high-quality sequential multilevel partitioner with advanced refinement techniques;
    \item \textbf{SHyPar}~\cite{sajadinia2025shypar}: an extension of KaHyPar that incorporates spectral information during coarsening; it runs 9 configurations of KaHyPar in parallel;
    \item \textbf{TritonPart}~\cite{bustany2023open}: a leading multi-threaded partitioner; we use its official implementation~\cite{triton} with default settings;
    \item \textbf{mtK-min}~\cite{gottesburen2024scalable}: a multi-threaded KaHyPar configuration (also named \textbf{mtKaHyPar}, 24 threads); we report the best cut and the total runtime over 100 runs;
    \item \textbf{mtK-mean}~\cite{gottesburen2024scalable}: the same configuration, reporting average cut size and runtime over 100 runs;
    \item \textbf{HySpecPro}~\cite{liang2026hyspecpro}: a leading GPU-accelerated spectral partitioner.
\end{itemize}

\begin{table}[!t]
\vspace{-2mm}
\caption{Bi-partitioning cut sizes on Titan23. The average improvement is normalized against hMETIS results.}
\vspace{-3mm}
\label{tab:sota20}
\centering
\scriptsize
\setlength{\tabcolsep}{2pt}
\renewcommand{\arraystretch}{1.05}
\begin{tabular}{l | r r r r r r r r r | r}
\toprule
\multirow{2}{*}{\textbf{Benchmark}} &
\multicolumn{10}{c}{\textbf{Cut Size at $\epsilon=2\%$}} \\
\cmidrule(lr){2-11}
& \rotatebox{90}{\textbf{hMETIS}}
& \rotatebox{90}{\textbf{SpecPart}}
& \rotatebox{90}{\textbf{MedPart}}
& \rotatebox{90}{\textbf{SHyPar}}
& \rotatebox{90}{\textbf{KaHyPar}}
& \rotatebox{90}{\textbf{TritonPart}}
& \rotatebox{90}{\textbf{mtK-min}}
& \rotatebox{90}{\textbf{mtK-mean}}
& \rotatebox{90}{\textbf{HySpecPro}}
& \rotatebox{90}{\textbf{FRESHPart}} \\
\midrule
sparcT1 core & 1066 & 1012 & 1067 & \textcolor{red}{974} & \textcolor{red}{974} & 1137 & 1099 & 1288 & 1091 & 1032 \\
neuron & 260 & 252 & 262 & \textcolor{red}{243} & 244 & 245 & 260 & 380 & 248 & 245 \\
stereo vision & 180 & 180 & 176 & \textcolor{red}{169} & \textcolor{red}{169} & 182 & 174 & 254 & 177 & 178 \\
des90 & 402 & 402 & 372 & 379 & 380 & 382 & 389 & 477 & 380 & \textcolor{red}{371} \\
SLAM spheric & \textcolor{red}{1061} & \textcolor{red}{1061} & \textcolor{red}{1061} & \textcolor{red}{1061} & \textcolor{red}{1061} & \textcolor{red}{1061} & \textcolor{red}{1061} & 1083 & \textcolor{red}{1061} & \textcolor{red}{1061} \\
cholesky mc & 285 & 285 & 283 & 283 & 283 & \textcolor{red}{282} & 382 & 756 & 283 & 285 \\
segmentation & 136 & 126 & 114 & \textcolor{red}{107} & \textcolor{red}{107} & \textcolor{red}{107} & 144 & 256 & 128 & \textcolor{red}{107} \\
bitonic mesh & 614 & 585 & 594 & 586 & 593 & 588 & 590 & 671 & 589 & \textcolor{red}{582} \\
dart & 844 & 807 & 805 & \textcolor{red}{784} & 924 & 846 & 817 & 1113 & 809 & 813 \\
openCV & 511 & 510 & 635 & 499 & 560 & \textcolor{red}{429} & 591 & 798 & 575 & \textcolor{red}{429} \\
stap qrd & 399 & 399 & 386 & \textcolor{red}{371} & \textcolor{red}{371} & 382 & 380 & 509 & 379 & 373 \\
minres & 215 & 215 & 215 & \textcolor{red}{207} & \textcolor{red}{207} & 208 & 252 & 351 & 210 & 210 \\
cholesky bdti & 1157 & \textcolor{red}{1156} & 1161 & \textcolor{red}{1156} & \textcolor{red}{1156} & 1197 & 1209 & 1792 & 1157 & \textcolor{red}{1156} \\
denoise & 722 & \textcolor{red}{416} & 516 & \textcolor{red}{416} & \textcolor{red}{416} & 479 & 456 & 533 & 439 & 419 \\
sparcT2 core & 1273 & 1244 & 1319 & \textcolor{red}{1183} & 1186 & 1310 & 1196 & 1301 & 1377 & 1251 \\
gsm switch & 5974 & 1827 & 1714 & 1621 & 1759 & \textcolor{red}{1496} & 2036 & 2789 & 1590 & 1706 \\
mes noc & 699 & \textcolor{red}{634} & 699 & 651 & 649 & 886 & 673 & 935 & 646 & 646 \\
LU230 & 4070 & \textcolor{red}{3273} & 3452 & 3602 & 4012 & 3426 & 3614 & 4390 & 3399 & 3298 \\
LU Network & 550 & 525 & 550 & \textcolor{red}{524} & \textcolor{red}{524} & 528 & 610 & 737 & 536 & \textcolor{red}{524} \\
sparcT1 chip2 & 1524 & 899 & 1129 & \textcolor{red}{873} & 874 & 1043 & 911 & 1196 & 903 & 878 \\
directrf & 646 & 574 & 646 & 632 & 646 & 576 & 662 & 731 & 506 & \textcolor{red}{498} \\
bitcoin miner & 1570 & \textcolor{red}{1297} & 1562 & 1514 & 1576 & 1489 & 1523 & 1582 & 1651 & 1518 \\
\midrule
\textbf{Impro.\ \% $\uparrow$}
& 0 & 10.99 & 6.82 & 12.16 & 9.95 & 8.90 & 4.16 & -28.00 & 9.42 & \textcolor{red}{13.09} \\
\bottomrule
\end{tabular}

\vspace{0.4em}

\begin{tabular}{l | r r r r r r r r r | r}
\toprule
\multicolumn{11}{c}{\textbf{Cut Size at $\epsilon=20\%$}} \\
\midrule
sparcT1 core & 1290 & 903 & 624 & 631 & 873 & 1016 & 1064 & 1210 & 682 & \textcolor{red}{589} \\
neuron & 270 & \textcolor{red}{206} & 270 & 244 & 244 & 245 & 251 & 360 & 216 & 211 \\
stereo vision & 143 & \textcolor{red}{91} & 93 & \textcolor{red}{91} & \textcolor{red}{91} & \textcolor{red}{91} & \textcolor{red}{91} & 103 & 97 & \textcolor{red}{91} \\
des90 & 441 & 358 & 349 & 345 & 380 & 353 & 382 & 481 & \textcolor{red}{345} & \textcolor{red}{345} \\
SLAM spheric & \textcolor{red}{1061} & \textcolor{red}{1061} & \textcolor{red}{1061} & \textcolor{red}{1061} & \textcolor{red}{1061} & \textcolor{red}{1061} & \textcolor{red}{1061} & 1076 & \textcolor{red}{1061} & \textcolor{red}{1061} \\
cholesky mc & 667 & 345 & \textcolor{red}{281} & 479 & 591 & \textcolor{red}{281} & \textcolor{red}{281} & 671 & \textcolor{red}{281} & \textcolor{red}{281} \\
segmentation & 141 & \textcolor{red}{78} & \textcolor{red}{78} & \textcolor{red}{78} & \textcolor{red}{78} & 80 & 86 & 148 & 82 & \textcolor{red}{78} \\
bitonic mesh & 590 & \textcolor{red}{483} & 493 & 506 & 493 & 506 & 588 & 675 & 489 & \textcolor{red}{483} \\
dart & 603 & 540 & 549 & \textcolor{red}{539} & 549 & 583 & 797 & 1127 & 667 & 598 \\
openCV & 554 & 518 & 554 & 473 & 554 & 451 & 625 & 778 & 534 & \textcolor{red}{428} \\
stap qrd & 295 & 295 & 287 & \textcolor{red}{275} & 287 & 281 & 364 & 490 & 276 & \textcolor{red}{275} \\
minres & 189 & 189 & \textcolor{red}{181} & 191 & \textcolor{red}{181} & 211 & 220 & 317 & 182 & \textcolor{red}{181} \\
cholesky bdti & 1024 & 947 & 1024 & \textcolor{red}{848} & 1024 & 861 & 1245 & 1680 & 858 & 857 \\
denoise & 478 & 224 & 224 & \textcolor{red}{220} & 224 & 231 & 340 & 436 & 284 & 226 \\
sparcT2 core & 1972 & 1245 & 1081 & \textcolor{red}{918} & 1081 & 1171 & 1192 & 1298 & 999 & 973 \\
gsm switch & 5352 & \textcolor{red}{1407} & 1503 & \textcolor{red}{1407} & 1503 & 1695 & 1641 & 2201 & 1434 & 1408 \\
mes noc & 633 & \textcolor{red}{617} & 633 & \textcolor{red}{617} & 633 & 623 & 634 & 860 & 631 & 620 \\
LU230 & 3276 & \textcolor{red}{2677} & 2720 & 2923 & 2720 & 3529 & 3586 & 4238 & 3017 & 2858 \\
LU Network & 528 & \textcolor{red}{524} & 528 & \textcolor{red}{524} & 528 & 529 & 526 & 532 & 542 & \textcolor{red}{524} \\
sparcT1 chip2 & 1029 & 783 & 877 & 757 & 877 & 939 & 847 & 1028 & 735 & \textcolor{red}{678} \\
directrf & 379 & \textcolor{red}{295} & 317 & \textcolor{red}{295} & 317 & \textcolor{red}{295} & 385 & 531 & 318 & 306 \\
bitcoin miner & 1255 & \textcolor{red}{1225} & 1255 & 1282 & 1255 & 1278 & 1232 & 1514 & 1231 & 1283 \\
\midrule
\textbf{Impro.\ \% $\uparrow$}
& 0 & 21.73 & 20.94 & 22.53 & 14.14 & 18.93 & 9.62 & -17.26 & 21.48 & \textcolor{red}{25.21} \\
\bottomrule
\end{tabular}
\vspace{-6mm}
\end{table}

\begin{table}[t]
\vspace{-4mm}
\caption{Ray concentration of the base, Geometry Feedback, and Search Feedback embeddings on the Titan23 benchmarks.}
\vspace{-4mm}
\label{tab:evolution-k12}
\centering
\scriptsize
\setlength{\tabcolsep}{2.5pt}
\begin{tabular}{l|r r|r r r r}
\toprule
\textbf{Benchmark} & $|V|$ & $|E|$ & Base Emb. & Geo. Emb. & Search Emb. & max of all \\
\midrule
sparcT1 core   & 91{,}976 & 92{,}827 & 0.951 & \textbf{0.992} & 0.968 & \textbf{0.992} \\
neuron         & 92{,}290 & 125{,}305 & 0.918 & \textbf{0.953} & 0.948 & \textbf{0.953} \\
stereo vision  & 94{,}050 & 127{,}085 & 0.936 & \textbf{0.981} & 0.931 & \textbf{0.981} \\
des90          & 111{,}221 & 139{,}557 & 0.853 & \textbf{0.946} & 0.921 & \textbf{0.946} \\
SLAM spheric   & 113{,}115 & 142{,}408 & 0.879 & \textbf{0.955} & 0.930 & \textbf{0.955} \\
cholesky mc    & 113{,}250 & 144{,}948 & 0.925 & \textbf{0.997} & 0.967 & \textbf{0.997} \\
segmentation   & 138{,}295 & 179{,}051 & 0.858 & 0.868 & \textbf{0.878} & \textbf{0.878} \\
bitonic mesh   & 192{,}064 & 235{,}328 & 0.857 & \textbf{0.921} & 0.868 & \textbf{0.921} \\
dart           & 202{,}354 & 223{,}301 & \textbf{0.992} & 0.967 & 0.937 & \textbf{0.992} \\
openCV         & 217{,}453 & 284{,}108 & 0.872 & 0.863 & \textbf{0.936} & \textbf{0.936} \\
stap qrd       & 240{,}240 & 290{,}123 & 0.948 & 0.983 & \textbf{0.984} & \textbf{0.984} \\
minres         & 261{,}359 & 320{,}540 & 0.903 & 0.923 & \textbf{0.955} & \textbf{0.955} \\
cholesky bdti  & 266{,}422 & 342{,}688 & 0.867 & \textbf{0.980} & 0.934 & \textbf{0.980} \\
denoise        & 275{,}638 & 356{,}848 & \textbf{0.978} & 0.886 & 0.972 & \textbf{0.978} \\
sparcT2 core   & 300{,}109 & 302{,}663 & 0.929 & \textbf{0.982} & 0.966 & \textbf{0.982} \\
gsm switch     & 493{,}260 & 507{,}821 & 0.976 & 0.836 & \textbf{0.990} & \textbf{0.990} \\
mes noc        & 547{,}544 & 577{,}664 & 0.826 & \textbf{0.864} & 0.854 & \textbf{0.864} \\
LU230          & 574{,}372 & 669{,}477 & 0.988 & \textbf{0.990} & 0.985 & \textbf{0.990} \\
LU Network     & 635{,}456 & 726{,}999 & 0.949 & \textbf{0.968} & 0.956 & \textbf{0.968} \\
sparcT1 chip2  & 820{,}886 & 821{,}274 & 0.952 & 0.930 & \textbf{0.958} & \textbf{0.958} \\
directrf       & 931{,}275 & 1{,}374{,}742 & \textbf{0.972} & 0.877 & 0.955 & \textbf{0.972} \\
bitcoin miner  & 1{,}089{,}284 & 1{,}448{,}151 & \textbf{0.962} & 0.883 & 0.956 & \textbf{0.962} \\
\midrule
\textbf{mean} & -- & -- & 0.922 & 0.934 & 0.943 & \textbf{0.961} \\
\bottomrule
\end{tabular}
\vspace{-2mm}
\end{table}

\begin{table}[!thb]
\vspace{-1mm}
\caption{Effect of ray-count selection at $\epsilon=20\%$. Cut-size improvement is measured relative to hMETIS.}
\vspace{-2mm}
\label{tab:k-selection}
\centering
\scriptsize
\setlength{\tabcolsep}{4pt}
\begin{tabular}{l|rrrrrr}
\toprule
\textbf{Ray count $K_r$} &  $4$ & $8$ & $12$ & $16$ & $24$ & $32$ \\
\midrule
Cut-size impr.\% $\uparrow$ & 24.31 & 24.42 & \textbf{25.21} & 24.83 & 24.52 & 23.66 \\
\bottomrule
\end{tabular}
\end{table}

\begin{figure*}[!t]
\centering
\vspace{-4mm}
\includegraphics[width=1.0\linewidth]{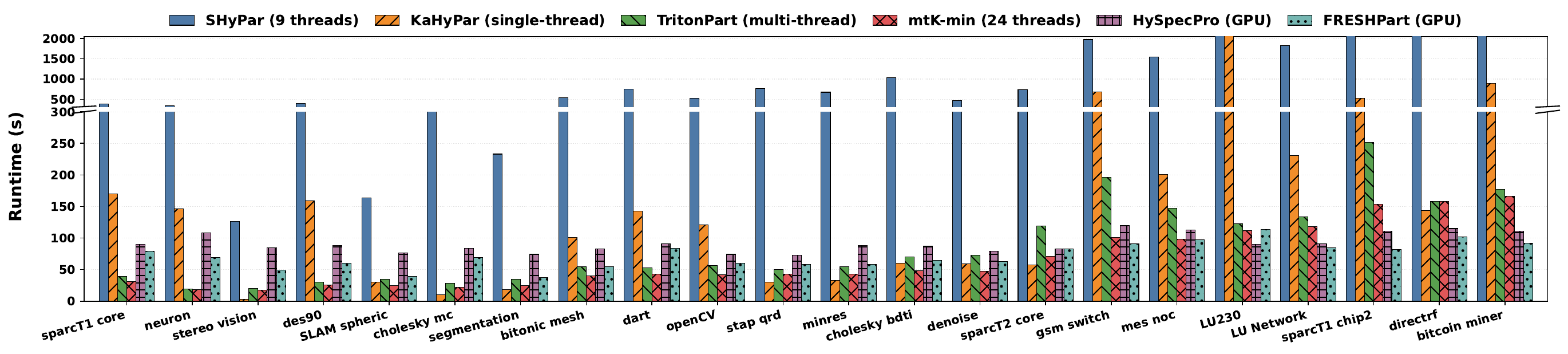}%
\vspace{-4mm}%
\caption{Runtime comparison among partitioners.}
\label{fig:runtime}
\end{figure*}

\vspace{-3.5mm}
\subsection{Partition Quality and Runtime}
\Cref{tab:sota20} summarizes the bi-partitioning cut sizes on Titan23. FRESHPart consistently outperforms both the leading multilevel and spectral partitioners, achieving the best reported cut sizes on 12 of the 22 benchmarks at $\epsilon=20\%$ and 8 of the 22 at $\epsilon=2\%$.

In terms of runtime, SpecPart reports an approximately $5\times$ runtime overhead over hMETIS~\cite{bustany2023k}, while MedPart requires up to two hours on the largest Titan benchmarks~\cite{liang2024medpart}. Detailed runtime comparisons among SHyPar, KaHyPar, TritonPart, mtKaHyPar, HySpecPro, and FRESHPart are shown in \Cref{fig:runtime}. FRESHPart is substantially faster than SHyPar, achieving more than an order-of-magnitude speedup on large benchmarks. Although FRESHPart is slower than KaHyPar, TritonPart, and mtKaHyPar on small hypergraphs, it consistently outperforms them on the largest benchmarks. Compared with the GPU-accelerated HySpecPro, FRESHPart achieves comparable runtime while consistently delivering higher partition quality. FRESHPart completes every Titan23 benchmark in under 120 s on a single GPU.

The runtime breakdown of FRESHPart shows that CMA-ES-based projection search dominates the runtime, accounting for approximately 85\% of the total execution time. The three Laplacian solves contribute about 10\%, while I/O together with the Geometry and Search Feedback computations account for only about 5\%.

\vspace{-2mm}
\subsection{Ablation Studies}
\begin{figure}[!t]
\centering
\vspace{-2mm}
\includegraphics[width=0.70\linewidth]{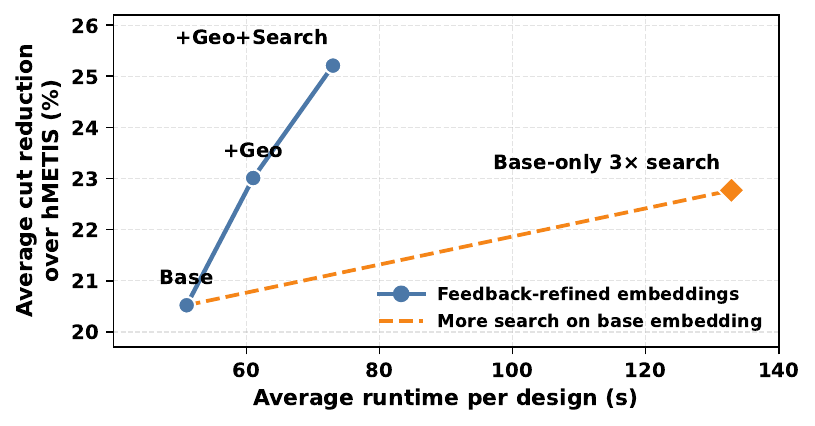}%
\vspace{-3mm}%
\caption{Partition quality versus runtime for FRESHPart variants.}
\vspace{-3mm}
\label{fig:tradeoff}
\end{figure}

\textbf{What Each Feedback Channel Contributes.}
As shown in \Cref{tab:evolution-k12}, both Geometry Feedback and Search Feedback increase the average ray concentration of the spectral embedding, producing embeddings better suited for projection-based partitioning. Correspondingly, \Cref{fig:tradeoff} shows that Geometry Feedback improves the average cut-size reduction over hMETIS by about 2\%, while Search Feedback provides a further 2\% improvement.

Introducing the Geometry and Search Feedback embeddings increases the runtime by roughly 50\% compared with searching only the base embedding. However, the resulting quality gain cannot be attributed simply to additional projection search. As shown in \Cref{fig:tradeoff}, allocating a large amount of search budget to the base embedding yields substantially smaller improvements. This indicates that the performance gains primarily arise from progressively refining the spectral embedding through the proposed feedback mechanisms, rather than from increased search effort alone.

\textbf{The Role of Adaptive Budget Allocation and Early Stopping.}
Our ablation study shows that disabling adaptive budget allocation and early stopping increases the average runtime by approximately 50\,s per design, while yielding virtually no improvement in partition quality. 

\textbf{The Impact of Ray Count Selection.}
The ray count $K_r$ affects both feedback stages, yet FRESHPart is
relatively robust to its choice, as shown in~\Cref{tab:k-selection}. Our smooth cross-ray weighting and mean
normalization may help limit sensitivity to $K_r$.

\textbf{Transfer to ISPD98 Benchmarks~\cite{tilos}.}
FRESHPart outperforms TritonPart by 4.69\% at 
$\epsilon=2\%$ and 7.06\% at $\epsilon=10\%$, respectively. Against HySpecPro, FRESHPart reduces mean cut by 6.00\% and 6.19\%, wins all 18 cases at both balance limits. FRESHPart is also faster than both methods. These results demonstrate that FRESHPart’s quality and efficiency transfer beyond Titan23.

\vspace{-2mm}
\section{Conclusion}
This paper revisits spectral hypergraph partitioning from the perspective of spectral embedding geometry. We show that the effectiveness of projection-based partitioning is closely tied to a ray-like embedding structure, and that graph structure, Laplacian design, and hyperedge reweighting all influence this geometry. Motivated by these insights, we propose FRESHPart, a feedback-driven spectral partitioning framework that refines spectral embeddings through Geometry Feedback and Search Feedback. Experiments on the Titan23 benchmark suite demonstrate that FRESHPart achieves SOTA partition quality while preserving runtime efficiency. We hope the proposed geometric perspective and feedback-driven framework will inspire further research on scalable embedding-aware hypergraph partitioning.



\bibliographystyle{ACM-Reference-Format}
\bibliography{ref,feng}

\end{document}